%% file: corl_2018.tex
\documentclass{article}

\usepackage[final]{corl_2018} 

\usepackage{graphicx}
\usepackage{subcaption}
\usepackage{amsmath}
\usepackage{bbm}
\usepackage{mathtools}
\usepackage{float}
\usepackage{hyperref}
\usepackage{multicol}
\usepackage{array,multirow}
\usepackage{adjustbox}
\usepackage{multicol}
\usepackage{tabularx}
\usepackage[utf8]{inputenc} 
\usepackage[T1]{fontenc}    
\usepackage{hyperref}       
\usepackage{url}            
\usepackage{booktabs}       
\usepackage{amsfonts}       
\usepackage{nicefrac}       
\usepackage{microtype}      
\usepackage{algorithm}
\usepackage{algorithmic}
\usepackage{wrapfig}
\usepackage{color, colortbl}

\usepackage{color, colortbl}
\definecolor{Gray}{gray}{0.9}

\usepackage{resizegather}

\title{Learning under Misspecified Objective Spaces} 

\author{
  Andreea Bobu \\
  UC Berkeley \\
  abobu@berkeley.edu
  \And
  Andrea Bajcsy  \\
  UC Berkeley \\
  abajcsy@berkeley.edu
  \And
  Jaime F. Fisac \\
  UC Berkeley \\
  jfisac@berkeley.edu
  \And
  Anca D. Dragan \\
  UC Berkeley \\
  anca@berkeley.edu
}
\usepackage[font=small]{caption}

\begin{document}
\maketitle


\vspace{-0.5cm}
\input{abstract}

\input{intro}

\input{method}

\input{experiments}

\input{conclusion}



\clearpage
\acknowledgments{This research is supported by the Air Force Office of Scientific Research (AFOSR) and the Open Philanthropy Project.}



\input{corl_2018.bbl}
\input{appendix}

\end{document}

%% file: abstract.tex
\begin{abstract}
Learning robot objective functions from human input has become increasingly important, but state-of-the-art techniques assume that the human's desired objective lies within the robot's hypothesis space. When this is not true, even methods that keep track of uncertainty over the objective fail because they reason about \emph{which} hypothesis might be correct, and not whether \emph{any} of the hypotheses are correct. We focus specifically on learning from physical human corrections during the robot's task execution, where not having a rich enough hypothesis space leads to the robot updating its objective in ways that the person did not actually intend. We observe that such corrections appear \emph{irrelevant} to the robot, because they are not the best way of achieving any of the candidate objectives. Instead of naively trusting and learning from every human interaction, we propose robots learn \emph{conservatively} by reasoning in real time about how relevant the human's correction is \emph{for the robot's hypothesis space}. We test our inference method in an experiment with human interaction data, and demonstrate that this alleviates unintended learning in an in-person user study with a robot manipulator.
\end{abstract}

%% file: intro.tex
\begin{figure}[h!]
\includegraphics[width=\textwidth]{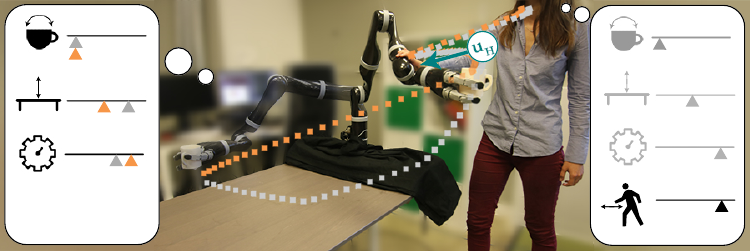}
\centering
\caption{The human pushes the robot to indicate she wants it to stay farther away from her, but this objective falls outside the robot's hypothesis space (i.e. robot does not reason about distance from humans). The robot detects a change in the distance to the table and erroneously learns to move closer to it (gray trajectory). Instead, with our method, the robot realizes that the human's correction does not make sense if what she wanted was a smaller distance to the table, and it does not erroneously learn from the physical interaction (orange trajectory).}
\label{fig:front_fig}
\end{figure}

\section{Introduction}
\label{intro}

Progress in robot motion planning and reinforcement learning has enabled faster and more efficient optimization of pre-specified objective functions. This progress on \emph{how} to solve optimization problems has opened up new questions about \emph{which} objective functions to use. New avenues of research focus on learning robot objectives from human input through demonstrations~\cite{abbeel2004apprenticeship,osa2018algorithmic}, teleoperation data~\cite{javdani2015shared}, corrections~\cite{jain2015learning,CORL}, comparisons~\cite{christiano2017preferences}, examples of what constitutes a goal~\cite{fu2018variational}, or even specified proxy objectives~\cite{HadfieldMenell2017InverseRD}. In this domain it is common to model the human as a utility-driven agent that chooses actions with respect to an internal objective function. Given measurements of the human's actions over time, the robot can attempt to solve an inverse reinforcement learning (IRL) problem \cite{Kalman1964inverse,Ng2000inverse}, by
running inference over a class of possible objective functions. By choosing a class of functions, these approaches implicitly assume that what the person wants (and is giving input about) lies inside of the robot's hypothesis space. Even when maintaining uncertainty over the objective (such as in Bayesian IRL~\cite{ramachandran2007bayesian}), state-of-the-art methods interpret human input as evidence about \emph{which} hypothesis is correct, rather than considering whether \emph{any} hypothesis is correct.

In this paper, we investigate the possibility that the robot does not have a rich enough hypothesis space to capture everything that the person might care about \cite{milli2017should}. 
It is important to stress that the model misspecification issue we are trying to mitigate is quite general and does not exclusively affect objectives based on hand-crafted features: any model, by definition, will ultimately fail to capture the underlying motivation of some human actions. While it may certainly be possible, and desirable, to continually increase the complexity of the robot's model to capture a richer space of possible objectives, there will still be a need to account for the presence of yet-unlearned components of the true objective. In this sense, our work is \emph{complementary} to open-world objective modeling efforts.

We focus our attention on a particular kind of human input: learning from physical human robot interaction (pHRI). In pHRI scenarios, a person works in close physical proximity with a robot and can apply forces on the robot as it is performing a task. Recent work \citep{CORL,HRI} has proposed methods to update the robot objective function from these interaction forces in real time in order to align the robot's performance with the human's preferences. For instance, if a human wants the robot to carry objects closer to the table, they push down on the manipulator and the robot updates its understanding of the task and moves closer to the table for the rest of the trajectory. However, if the robot's objective space does \textit{not} include features that encode other objectives the human may care about (e.g. distance to the human's body), then  the robot can mistakenly learn about other features it \textit{does} know about (like moving closer to the table---gray trajectory in Fig.~\ref{fig:front_fig}). In other words, even if the human's correction is not relevant to the robot's hypothesis space, the robot will mistakenly learn anyway.

Our goal is to detect whether human input is \textit{relevant}, i.e. meant to correct something known to the robot. We propose that when a correction is relevant, it is informative \textit{while also minimizing human effort}. If there are easier ways to achieve the same change in the objective, then the observed change is likely not what the person intended---they meant to correct something else (or possibly nothing at all). In doing this, we take inspiration from prior work that has used apparent rationality to adjust confidence in predictions about human behavior \citep{fisac2018probabilistically}, but introduce a method for detecting correction relevance in real time for continuous and high-dimensional state and action spaces. We test this method in an experiment with human physical interaction data on a robotic manipulator, and test its implications for learning in a user study.

%% file: method.tex
\section{Method}
\label{method}


We consider settings where a robot is collaborating with a person to perform everyday tasks by learning from pHRI, but their feature spaces might not be shared. When the robot receives human input, it should be able to reason about the relevance of the action and replan its trajectory accordingly.

\noindent\textbf{A Graphical Model for Correction Relevance.} Let $x$ describe the robot's state (joint position and velocity) and $u_R$ the robot's action (the control torque it applies). By default, the robot generates actions via an impedance controller to track a trajectory $\xi_R = x^{0:T} \in \Xi$. This trajectory is optimal with respect to the robot's objective which is encoded as a linear combination of features of the state, $\phi(x^t)$ \cite{abbeel2004apprenticeship,maxent}. The robot might adapt its objective over time, which means it replans and tracks a new trajectory corresponding to its understanding of the objective at the current time $t$ by solving $\xi_R^t = \arg\min_{\xi} {\hat\theta}^\top\Phi(\xi)$. Here  $\Phi(\xi) = \sum_{x^{\tau}\in\xi} \phi(x^{\tau})$ is the total feature count along a particular trajectory $\xi$, and $\hat\theta \in \Theta$ represents the current weights on these features. 

At any timestep during the trajectory execution, the human may physically interact with the robot, inducing a joint torque, $u_H$. Following \citep{von1945theory,baker2007goal,CORL}, we model the human as a noisily-rational agent that chooses actions which approximately minimize the robot's cost:
\begin{equation}
P(u_H \mid \xi_R; \beta, \theta) = \frac{e^{-\beta ({\theta}^\top\Phi(\xi_H) + \lambda\|u_H\|^2)}}{\int e^{-\beta ({\theta}^\top\Phi(\bar{\xi}_H) + \lambda\|\bar{u}_H\|^2)}d\bar{u}_H}\enspace.
\label{eq:boltzmann}
\end{equation}
In this model, $\beta \in [0,\infty)$ is the rationality coefficient that determines the variance of the distribution over human actions, and $\xi_H$ is the trajectory induced by the human's action $u_H$. The coefficient $\beta$ adjusts how noisily the human picks $u_H$. When the person's correction is relevant, $\beta$ is large, while during irrelevant corrections $\beta$ is small, resulting in seemingly random behavior. Lastly, in this work, we compute $\xi_H$ by deforming the robot's current trajectory $\xi_R$ in the direction of $u_H$ via $\xi_H = \xi_R + \mu A^{-1}U_H$ \citep{CORL}. Here, $\mu>0$ scales the magnitude of the deformation, $A$ defines a norm on the Hilbert space of trajectories and dictates the deformation shape \citep{deformation}, and $U_H=u_H$ at the current time and $U_H=0$ otherwise\footnote{We used a norm $A$ based on acceleration, consistent with \citep{CORL}, but other norm choices are possible as well.}. When choosing actions, the human is modeled as trading off between inducing a good trajectory $\xi_H$ with respect to~$\theta$, and minimizing their effort. 

Ideally, with this model of human actions we would perform inference over both the rationality coefficient $\beta$ and the parameters $\theta$ by maintaining a joint Bayesian belief $b(\beta, \theta)$. Our probability distribution over $\theta$ would automatically adjust for relevant corrections, whereas for irrelevant ones the robot would stay closer to its prior on $\theta$: the robot would implicitly realize that there is a better $u_H$ for each possible $\theta$, and therefore no $\theta$ explains the correction. Unfortunately, this is not possible in real time given the continuous and possibly high-dimensional nature of the weight-space, $\Theta$.

To alleviate the computational challenge of performing joint inference over $\beta$ and $\theta$, we introduce a binary variable $r \in \{0,1\}$ which (a) indicates interaction relevance, and (b) allows us to split inference over $\beta$ and $\theta$ into two separate computations. 
When $r=1$, the interaction appears less noisy (i.e. $\beta$ is large) and the human's desired modification of the robot's behavior can be well explained by \emph{some} feature weight vector $\theta \in \Theta$. As a result, the interaction $u_H$ is likely to be efficient for the cost encoded by this $\theta$. Conversely, when $r=0$, the interaction appears noisy (i.e. $\beta$ is small) and the person is correcting some feature(s) that the robot does not have access to. Thus, the interaction $u_H$ is substantially less likely to be efficient for any cost expressible within the space $\Theta$.

\begin{figure}[t!]
\centering
\includegraphics[width=0.85\textwidth]{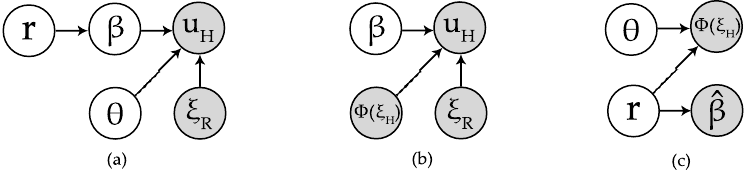}
\caption{(a) In the complete graphical model, $u_H$ is an observation of the true objective ($\theta$) and rationality coefficient ($\beta$);
inference over $\theta$ is intractable.
(b) We use the proxy variable $\Phi(\xi_H)$ to first estimate $\beta$. 
(c) We interpret the estimate $\hat{\beta}$ as an indirect observation of the unobserved $r$, which we then use for the $\theta$ estimate.}
\label{fig:graphical_models}
\end{figure}

Overall, this forms the graphical model in \ref{fig:graphical_models}a, with $P(u_H \mid \xi_R; \beta, \theta)$ defined above, and the distribution $P(\beta \mid r)$ (describing what $\beta$ we expect under relevant or irrelevant corrections) to be learned empirically. Next, we simplify the problem to enable real time computation while preserving the benefit of automatically adjusting the learning depending on relevance. 

\noindent\textbf{Approximate Inference.}

We simplify inference by leveraging a proxy variable $\Phi(\xi_H)$---the feature values of the corrected trajectory $\xi_H$---to split the problem into two stages. Rather than looking at how relevant $u_H$ is overall, we estimate rationality $\beta$ (and therefore relevance $r$) based on how efficiently $u_H$ produced the feature values of $\Phi(\xi_H)$. If there is a more efficient correction that would induce $\Phi(\xi_H)$, this suggests that the correction was not relevant to the robot's hypothesis space. Then, armed with an estimate of relevance, we infer $\theta$ from $\Phi(\xi_H)$. 

These two steps are captured by the two simplified graphical models in Figure \ref{fig:graphical_models}b and Figure \ref{fig:graphical_models}c. The first relates $u_H$ to $\beta$, $\xi_R$, and the induced feature values $\Phi(\xi_H)$ by modeling that $u_H$ should achieve the $\Phi(\xi_H)$ with minimal effort. The conditional probability in this Bayes net is: 

%
%
\vspace{-0.5cm}
\begin{equation}
	P(u_H \mid \beta, \Phi(\xi_H), \xi_R) = \frac{e^{-\beta( \|u_H\|^2)}}{\int e^{-\beta( \|\bar{u}_H\|^2+ \lambda\|\Phi(\bar{\xi}_H) - \Phi(\xi_H)\|^2 )}d\bar{u}_H}\enspace,\label{eq:bayesnetb_model}
\end{equation}
where the cost function is $\beta( \|u_H\|^2 + \lambda\|\Phi(\xi_H) - \Phi(\xi_H)\|^2)$, with $\lambda\|\Phi(\bar{\xi}_H) - \Phi(\xi_H)\|^2 $ a strong penalty on deviating from the induced features (this penalty is 0 in the numerator since the observed $\Phi(\xi_H)$ is exactly the one induced by $u_H$). This will enable us to estimate a $\hat{\beta}$ from the observed $u_H$.

The second simplified graphical model (Figure \ref{fig:graphical_models}c) relates the induced feature values $\Phi(\xi_H)$ to $\theta$ as a function of the relevance. When $r=1$, the induced features should have low cost with respect to $\theta$; when $r=0$, the induced features \emph{do not depend on $\theta$}, and we model them as Gaussian noise centered at the current robot trajectory feature values:
\begin{equation}
P(\Phi(\xi_H) \mid \theta, r) = \begin{cases} \frac{e^{-\theta^\top\Phi(\xi_H)}}{\int e^{-\theta^{\top}\Phi(\tilde{\xi}_H)} d\tilde{\xi}_H} ,& r=1 \\ \left(\frac{\lambda}{\pi}\right)^\frac{k}{2} e^{-\lambda ||\Phi(\xi_H)-\Phi(\xi_R)||^2} ,& r=0 \end{cases}
\label{eq:bayesnetc_model}
\end{equation}
with the constant in the $r=0$ case corresponding to the normalization term of the normal distribution.

Finally, this graphical model also relates the estimated $\hat{\beta}$ from the first simplified model to $r$ by a $P(\hat{\beta} \mid r)$. We fit this distribution from controlled user interaction samples where we know $r$. For each sample interaction, we compute a $\hat{\beta}$ (using Equation \ref{eq:beta_map} below) and label it with the corresponding binary $r$ value. We fit a chi-squared distribution to these samples to obtain the probability distributions for $P(\hat{\beta} \mid r=0)$ and $P(\hat{\beta} \mid r=1)$. The resulting distribution fits are shown in Figure \ref{fig:histograms}.

\noindent\textbf{Estimating $\beta$.} The integral in Equation \ref{eq:bayesnetb_model} is 
computationally prohibitive, 
but we can approximate it via a Laplace approximation (detailed in Appendix \ref{app:laplace}): 
\begin{equation}
	P(u_H \mid \beta, \Phi(\xi_H), \xi_R) \approx \frac{e^{-\beta(  \|u_H\|^2)}}{ e^{-\beta( \|{u_H^*}\|^2 + \lambda\|\Phi({\xi_H^*}) - \Phi(\xi_H)\|^2)}}\sqrt{\frac{\beta^k|H_{u^*_H}|}{2\pi^k}}
\enspace,
\label{eq:laplace}
\end{equation}
where $k$ is the action space dimensionality and $H_{u_H^*}$ is the Hessian of the cost function around $u_H^*$. We obtain the optimal action $u_H^*$ by solving the constrained optimization problem:
\begin{equation}
\begin{aligned}
& \underset{u}{\text{minimize}}
& & \|u\|^2 \\
& \text{subject to}
& & \Phi(\xi_R+\mu A^{-1}U) - \Phi(\xi_H) = 0.
\end{aligned}
\label{opt:optimal_uH}
\end{equation}
Notice that by formulating the problem as a constrained optimization, the $\lambda$-term in Equation \ref{eq:laplace} is close to 0 and vanishes, so the final conditional probability becomes:
\begin{equation}
	P(u_H \mid \beta, \Phi(\xi_H), \xi_R) = e^{-\beta(\|u_H\|^2-\|u_H^*\|^2 )}\sqrt{\frac{\beta^k|H_{u^*_H}|}{2\pi^k}}\enspace.
\label{eq:laplace2}
\end{equation}
We can obtain an estimate of the rationality coefficient via a maximum likelihood estimate:
\begin{gather}
	\hat{\beta} = \arg\max_{\beta} \{\log(P(u_H \mid \beta, \Phi(\xi_H), \xi_R))\} = \arg\max_{\beta} \{-\beta(\|u_H\|^2-\|u_H^*\|^2 ) + \log(\sqrt{\frac{\beta^k|H_{u^*_H}|}{2\pi^k}})\}
\end{gather}
and setting the gradient to zero:
\begin{equation}
	\hat{\beta} = \frac{k}{2(\|u_H\|^2-\|u_H^*\|^2)}\enspace.
\label{eq:beta_map}
\end{equation}
\noindent\textbf{Intuitive Interpretation of $\hat{\beta}$.} The estimator above yields a high value when the difference between $u_H$ and $u_H^*$ is small, i.e. the person's correction achieves the induced features $\Phi(\xi_H)$ efficiently. For instance, if $\xi_H$ brings the robot closer to the table, and $u_H$ pushes the robot straight towards the table, $u_H$ is an efficient way to induce those new feature values. However, when there is a much more efficient alternative (e.g. when the person pushes mostly sideways rather than straight towards the table), $\hat{\beta}$ will be small. Efficient ways to induce the feature values will suggest relevance, inefficient ones will suggest irrelevance (i.e. there is something else that explains the correction).


\noindent\textbf{Estimating ${\theta}$.}
Lastly, we can infer a new $\theta$ estimate based on the model from Figure  \ref{fig:graphical_models}c.
We do this tractably by interpreting the estimate $\hat\beta$ obtained from \eqref{eq:beta_map} as an indirect observation of the unknown variable $r$.
We empirically characterize the conditional distribution $P(\hat\beta\mid r)$, and then combine the learned likelihood model with an initial uniform prior $P(r)$ to maintain a Bayesian posterior on $r$ based on the evidence $\hat\beta$ constructed from human observations at deployment time,
$P(r \mid \hat\beta) \propto P(\hat\beta \mid r) P(r)$.
%
Further, since we wish to obtain a posterior estimate of the human's objective $\theta$, we use the model from Figure  \ref{fig:graphical_models}c to obtain the posterior probability measure
\begin{equation}\label{eq:theta_from_beta_hat}
P(\theta \mid \Phi(\xi_H),\hat\beta) \propto \sum_{r \in \{0,1\} } P\big(\Phi(\xi_H) \mid \theta, r\big) P(r \mid \hat\beta) P(\theta)
\enspace.
\end{equation}
%
Following \cite{CORL}, we approximate the partition function in the human's policy  \eqref{eq:bayesnetc_model} by the exponentiated cost of the robot's original trajectory, thereby obtaining
\begin{equation}
P(\Phi(\xi_H) \mid \theta, r=1) = 
	e^{-\theta^\top\big(\Phi(\xi_H)-\Phi(\xi_R)\big)}\enspace. 
\label{eq:bayesnetc_with_constants}
\end{equation}
We also consider a Gaussian prior distribution of $\theta$ around the robot's current estimate $\hat\theta$:
\begin{equation}
P(\theta) = \frac{1}{(2\pi\alpha)^\frac{k}{2}} e^{-\frac{1}{2\alpha} ||\theta-\hat{\theta}||^2}
\enspace.
\end{equation}
The maximum-a-posteriori estimate of the human's objective $\theta$ is therefore the solution maximizer of
\begin{gather}\begin{aligned}\label{eq:posterior_theta}
P(\theta&) \Big[ P(r=1 \mid \hat\beta)P\big(\Phi(\xi_H)\mid\theta,r=1\big) +
	 P(r=0 \mid \hat\beta)P\big(\Phi(\xi_H)\mid \theta,r=0\big) \Big] \\
     =& \frac{1}{(2\pi\alpha)^\frac{k}{2}}
e^{-\frac{1}{2\alpha}||\theta-\hat{\theta}||^2}
   \Big[ P(r=1 \mid \hat\beta) e^{-\theta^\top\big(\Phi(\xi_H)-\Phi(\xi_R)\big)} + P(r=0 \mid \hat\beta) \left(\frac{\lambda}{\pi}\right)^\frac{k}{2} e^{-\lambda||\Phi(\xi_H)-\Phi(\xi_R)||^2}\Big]
\enspace.
\end{aligned}\end{gather}
%
Differentiating \eqref{eq:posterior_theta} with respect to $\theta$ and equating to 0 gives
the maximum-a-posteriori update rule:
\begin{gather}\label{eq:theta_update}
\hat\theta' = \hat\theta - \alpha \frac{P(r=1 \mid \hat\beta) e^{-\hat\theta'^\top\big(\Phi(\xi_H)-\Phi(\xi_R)\big)}}
{P(r=1 \mid \hat\beta) e^{-\hat\theta'^\top\big(\Phi(\xi_H)-\Phi(\xi_R)\big)} +
P(r=0 \mid \hat\beta) \left(\frac{\lambda}{\pi}\right)^\frac{k}{2} e^{-\lambda||\Phi(\xi_H)-\Phi(\xi_R)||^2}}
\big(\Phi(\xi_H)-\Phi(\xi_R)\big)
\enspace.
\end{gather}
We note that due to the coupling in $\hat\theta'$, the solution to \eqref{eq:theta_update} is non-analytic and can instead be obtained via classical numerical approaches like Newton-Raphson or quasi-Newton methods.

In previous objective-learning approaches including \cite{CORL} and \cite{maxent}, it is implicitly assumed that all human actions are fully explainable by the robot's representation of the objective function space ($r=1$), leading to the simplified update
\begin{equation}\label{eq:corl_update}
\hat\theta' = \hat\theta - \alpha
\big(\Phi(\xi_H)-\Phi(\xi_R)\big)
\enspace,
\end{equation}
which can be easily seen to be a special case of \eqref{eq:theta_update} when $P(r=0 \mid \hat\beta)\equiv 0$.
Our proposed update rule therefore \emph{generalizes} commonly-used objective-learning formulations to cases where the human's underlying objective function is not fully captured by the robot's model.
We expect that this extended formulation will enable learning that is more robust to misspecified or incomplete human objective parameterizations. Once we obtain the $\hat{\theta}'$ update, we replan the robot trajectory in its 7-DOF configuration space with an off-the-shelf trajectory optimizer\footnote{We provide an open-source implementation of our method \href{https://github.com/andreea7b/beta_adaptive_pHRI}{here}. We used TrajOpt \citep{trajopt}, which is based on sequential quadratic programming and uses convex-convex collision checking.}. 

\noindent\textbf{Intuitive Interpretation of $\hat{\theta}$.} The update rule changes the weights in the objective in the direction of the feature difference as well, but how much it does so depends on the probability assigned to the correction being relevant. At one extreme, if we know with full certainty that the correction is relevant, then we do the full update as in traditional objective learning. But crucially, at the other extreme, if we know that the correction is irrelevant, we do not update at all and keep our prior belief.

%% file: experiments.tex
\section{Offline $\beta$-estimation Experiments}
    
\begin{figure*}
\centering
  \centering
  \includegraphics[scale=.38]{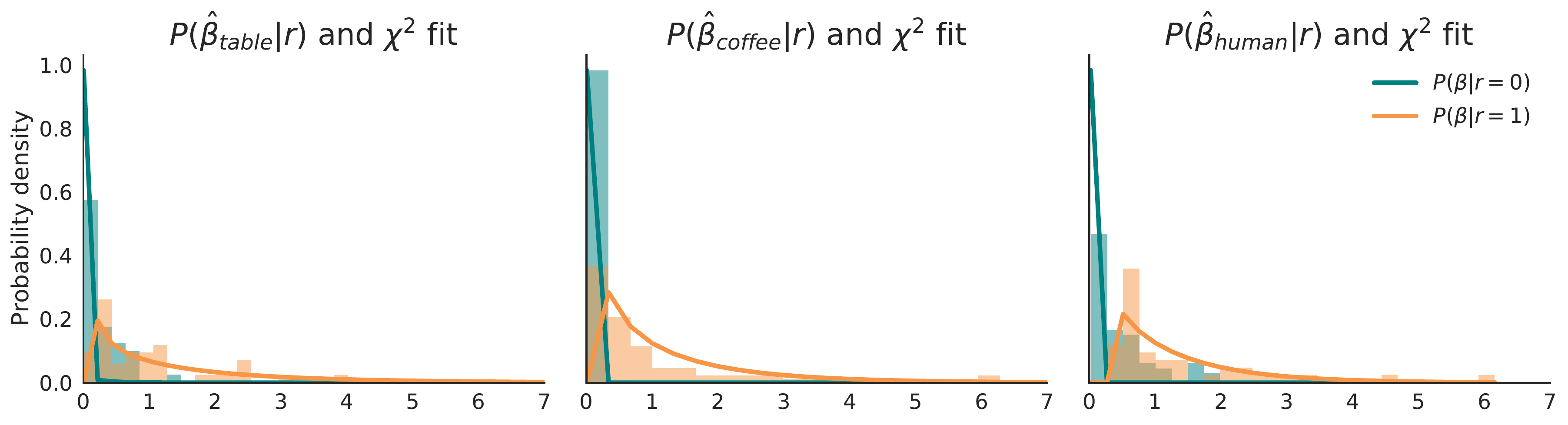}
  \caption{Empirical estimates for $P(\hat{\beta} \mid r)$ and their corresponding chi-squared ($\chi^2$) fits.}
  \label{fig:histograms}
\end{figure*}

We evaluate our rationality estimates with \emph{human data} collected offline.

\noindent\textbf{Independent and Dependent Variables.} We recruited 12 people to physically interact with a JACO 7DOF robotic arm as it moves from a start to a goal pose while holding a cup. We asked the participants to intentionally correct a feature (that the robot may or may not know about): adjusting the distance of the end effector from the table, adjusting the distance from the person, or adjusting the cup's orientation\footnote{Because users tend to accidentally correct more than one feature, we perform $\beta$-inference separately for each feature. This requires more overall computation (although still linear in the number of features and can be parallelized) and a separate $P(\hat{\beta} \mid r)$ estimate for each feature.}. 
In this data collection stage, the robot did not perform learning and simply tracked a predefined trajectory with impedance control \citep{impedance}. The participants were instructed to intervene only once during the robot's task execution, in order to record a single physical correction. 

Next, we ran our inference algorithm using the recorded human interaction torques and robot joint angle information. We measured what $\hat{\beta}$ would have been for each interaction if the robot knew about a given subset of the features. By changing the subset of features for the robot, we changed whether any given human interaction was \emph{relevant} to the robot's hypothesis space.

\noindent\textbf{Hypotheses:}

\noindent\textbf{H1.} \textit{Relevant interactions result in high $\hat{\beta}$, whereas interactions that change a feature the robot \textbf{does not} know about result in low $\hat{\beta}$ for all features the robot \textbf{does} know about.}
\vspace{-1mm}

\noindent\textbf{H2.} \textit{Not reasoning about relevant interactions and, instead, indiscriminately learning from every update leads to significant unintended learning.}

\noindent\textbf{Analysis.} 
We ran a repeated-measures ANOVA to test the effect of relevance on our $\hat{\beta}$. We found a significant effect ($F(1,521)=9.9093$, $p=0.0017$): when the person was providing a relevant correction, $\hat{\beta}$ was significantly higher. This supports our hypothesis H1.

\begin{figure}[t!]
\begin{subfigure}[b]{0.5\textwidth}
\centering
\includegraphics[height=.5\textwidth]	{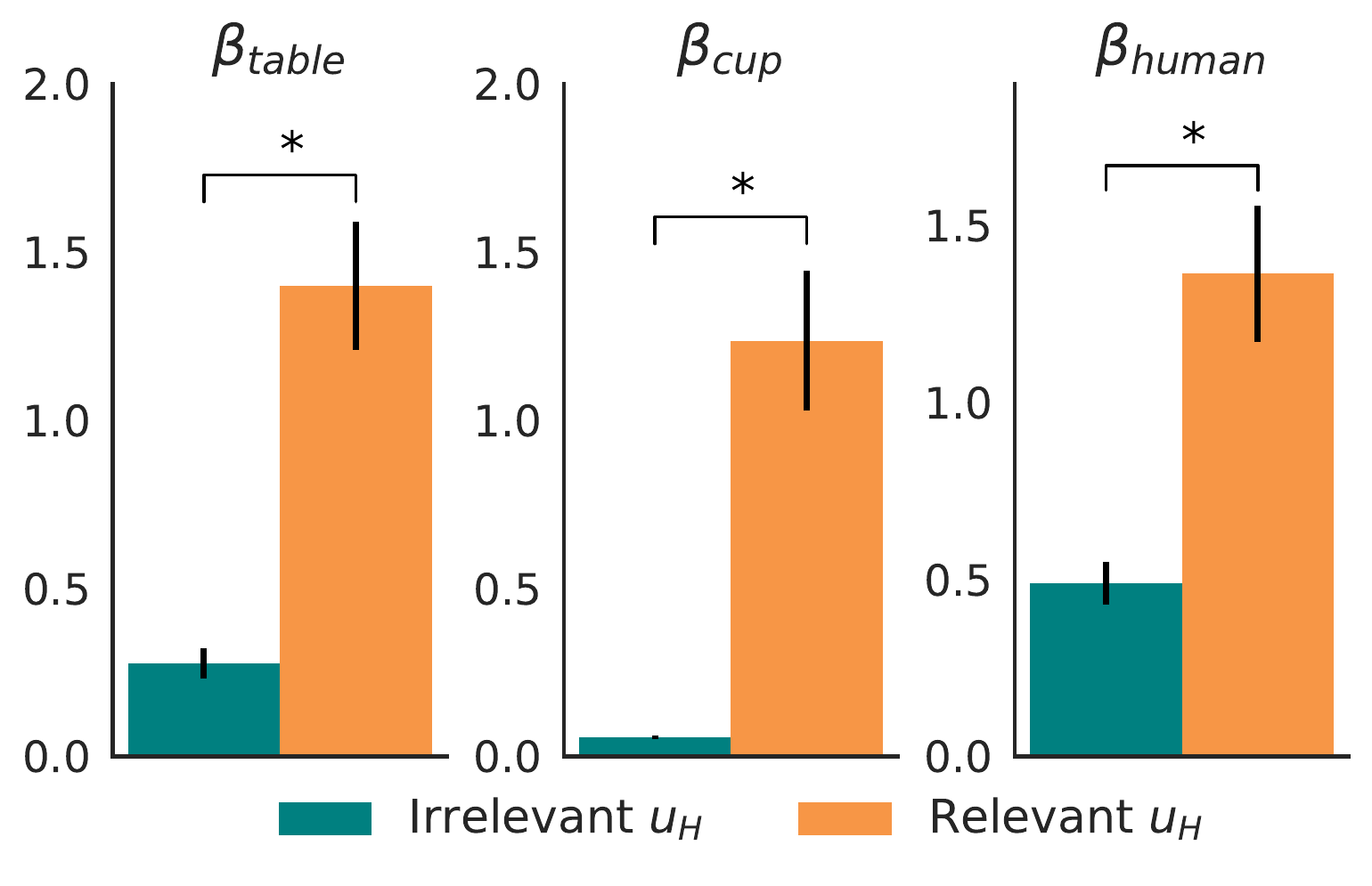}\\
\caption{Average $\beta$ for relevant and irrelevant interactions.}
 \label{fig:beta_relevance}
\end{subfigure}
\begin{subfigure}[b]{0.5\textwidth}
\centering
\includegraphics[height=.5\textwidth]{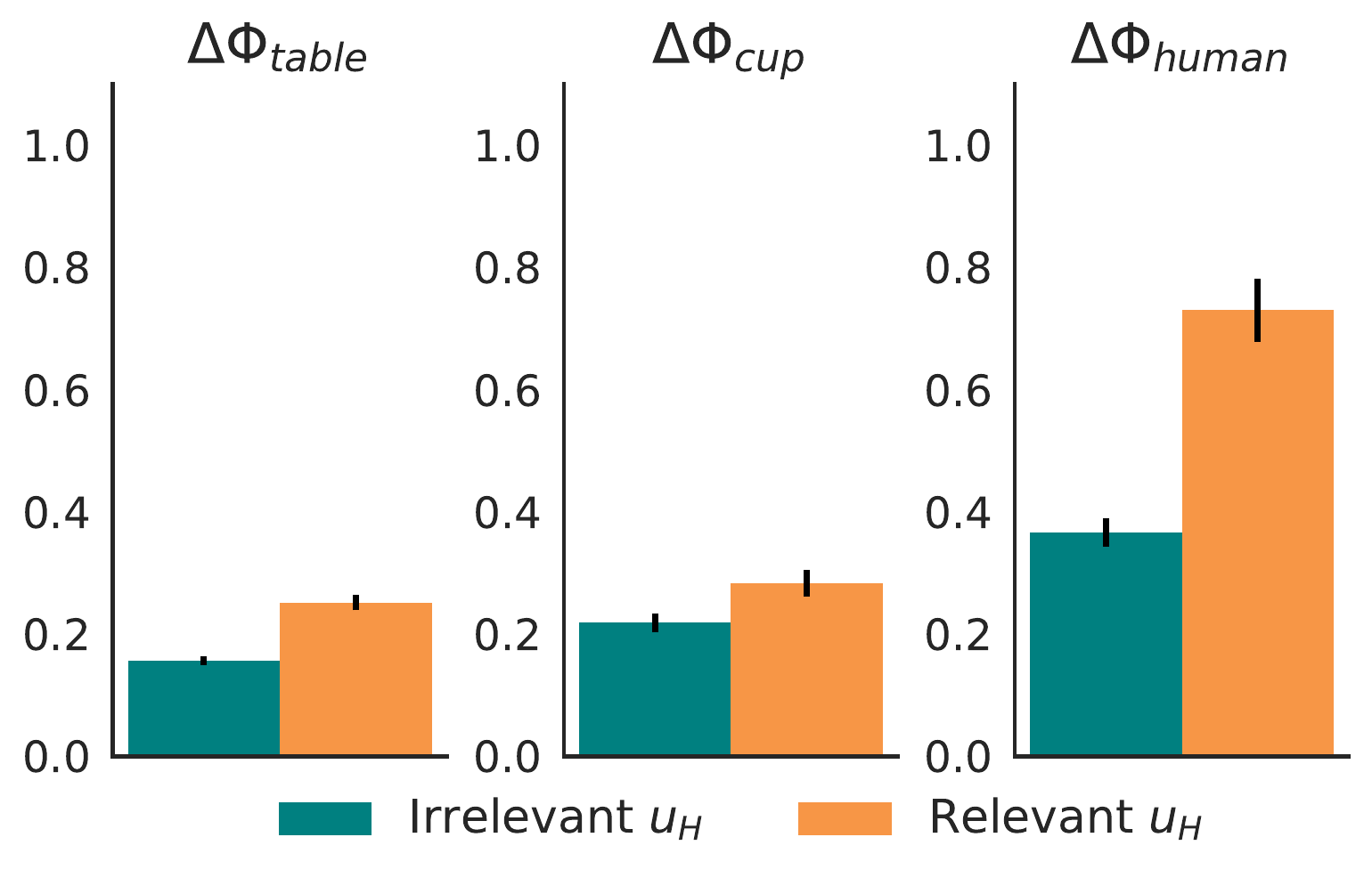}\\
\caption{Average $\Delta\Phi$ for relevant and irrelevant interactions.}
 \label{fig:update_relevance}
\end{subfigure}
\vspace{1mm}
\caption{$\beta$ values are significantly larger for relevant actions than for irrelevant ones. Feature updates are non-negligible even during irrelevant actions, which leads to significant unintended learning for fixed-$\beta$ methods.}
\label{fig:offline_relevant}
\end{figure}

Figure \ref{fig:beta_relevance} plots $\hat{\beta}$ under the relevant (orange) and irrelevant (blue) conditions. Whereas the irrelevant interactions end up with $\hat{\beta}$s close to 0, relevant corrections have higher mean and take on a wider range of values, reflecting varying degrees of human performance in correcting something the robot knows about. We fit per-feature chi-squared distributions for $P(\hat{\beta} \mid r)$ for each value of $r$ which we will use to infer $r$ and, thus, $\theta$ online. In addition, Figure \ref{fig:update_relevance} illustrates that even for irrelevant human actions $u_H$, the resulting feature difference $\Delta\Phi = \Phi(\xi_H)-\Phi(\xi_R)$ is non-negligible. This supports our second hypothesis, H2, that not reasoning about action relevance is detrimental to learning performance when the robot receives misspecified updates.

\section{Online Learning User Study}

Our offline experiments suggested that $\hat{\beta}$ can be used as a measure  of whether physical interactions are relevant and should be learned from. Next, we conducted an IRB-approved user study to investigate the implications of using these estimates during learning. During each experimental task, the robot began with a number of incorrect weights and participants were asked to physically correct the robot. Locations of the objects and human were kept consistent in our experiments across tasks and users to control for confounds\footnote{We assume full observability of where the objects and the human are, as the focus of this paper is not sensing.}. The planning and inference were done for robot trajectories in 7-dimensional configuration space, accounting for all relevant constraints including joint limits and self-collisions, as well as collisions between obstacles in the workspace and any part of the robot’s body.\footnote{For video footage of the experiment, see: \href{https://youtu.be/stnFye8HdcU}{https://youtu.be/stnFye8HdcU}} 

\subsection{Experiment Design}

\noindent\textbf{Independent Variables.}
We used a 2 by 2 factoral design. We manipulated the pHRI learning strategy with two levels (fixed and adaptive learning), and also relevance (human corrected for features inside or outside the robot's hypothesis space). In the fixed learning strategy, the robot updated its feature weights from a given interaction via \eqref{eq:corl_update}. In the adaptive learning strategy, the robot updates its feature weights via \eqref{eq:theta_update}. The offline experiments above provided us an estimate for $P(r\mid \hat{\beta})$ that we used in the gradient update. 

\noindent\textbf{Dependent Measures.}

\textbf{\emph{Objective.}} To analyze the objective performance of the two learning strategies, we focused on comparing two main measurements: the length of the $\hat{\theta}$ path through weight space as a measurement of the learning process, and the regret in feature space measured by
$|\Phi(\xi_{\theta^*}) - \Phi(\xi_{actual})|$.

\textbf{\emph{Subjective.}} For each condition, we administered a 7-point Likert scale survey about the participant's interaction experience (Table \ref{tab:likert}). We separate the survey into 3 scales: task completion, task understanding, and unintended learning.

\noindent\textbf{Hypotheses:}

\textbf{H1.} \textit{On tasks where humans try to correct \textbf{inside} the robot's hypothesis space (relevant corrections), detecting irrationality is not inferior to assuming rational human behavior.}
\vspace{-1mm}

\textbf{H2.} \textit{On tasks where humans try to correct \textbf{outside} the robot's hypothesis space (irrelevant corrections), detecting irrationality reduces unintended learning.}
\vspace{-1mm}

\textbf{H3.} \textit{On tasks where they tried to correct \textbf{inside} the robot's hypothesis space, participants felt like the two methods performed the same.}
\vspace{-1mm}

\textbf{H4.} \textit{On tasks where they tried to correct \textbf{outside} the robot's hypothesis space, participants felt like our method reduced unintended learning.}

\noindent\textbf{Tasks.}
We designed four experimental household manipulation tasks for the robot to perform in a shared workspace. For each experimental task, the robot carried a cup from a start to end pose with an initially incorrect objective. Participants were instructed to physically intervene to correct the robot's behavior during the task. Similar to state-of-the-art methods, all the features were chosen to be intuitive to a human to ensure that participants could understand how to correct the robot.

In Tasks 1 and 2, the robot's default trajectory took a cup from the participant and put it down on the table, but carried the cup too high above the table. In Tasks 3 and 4, the robot also took a cup from the human and placed it on the table, but this time it initially grasped the cup at the wrong angle, requiring human assistance to correct end-effector orientation to an upright position. For Tasks 1 and 3, the robot knew about the feature the human was asked to correct for ($r=1$) and participants were told that the robot should be compliant. For Tasks 2 and 4, the correction was irrelevant ($r=0$) and participants were instructed to correct any additional unwanted changes in the trajectory.

\noindent\textbf{Participants.}
We used a within-subjects design and randomized the order of the learning methods during experiments. In total, we recruited 12 participants (6 females, 6 males, aged 18-30) from the campus community, 10 of which had technical backgrounds and 2 of which did not. None of the participants had experience interacting with the robot used in our experiments.

\noindent\textbf{Procedure.}
Every participant was assigned a random ordering of the two methods, and performed each task without knowing how the methods work. At the beginning of each task, the participant was shown the incorrect default trajectory, followed by the desired trajectory. Then the participant performed a familiarization round, followed by two recorded experimental rounds. After answering the survey, the participant repeated the procedure for the other method. 

\vspace{-0.3cm}

\begin{figure}[t!]
\begin{subfigure}[b]{0.5\textwidth}
\centering
\includegraphics[height=.5\textwidth]	{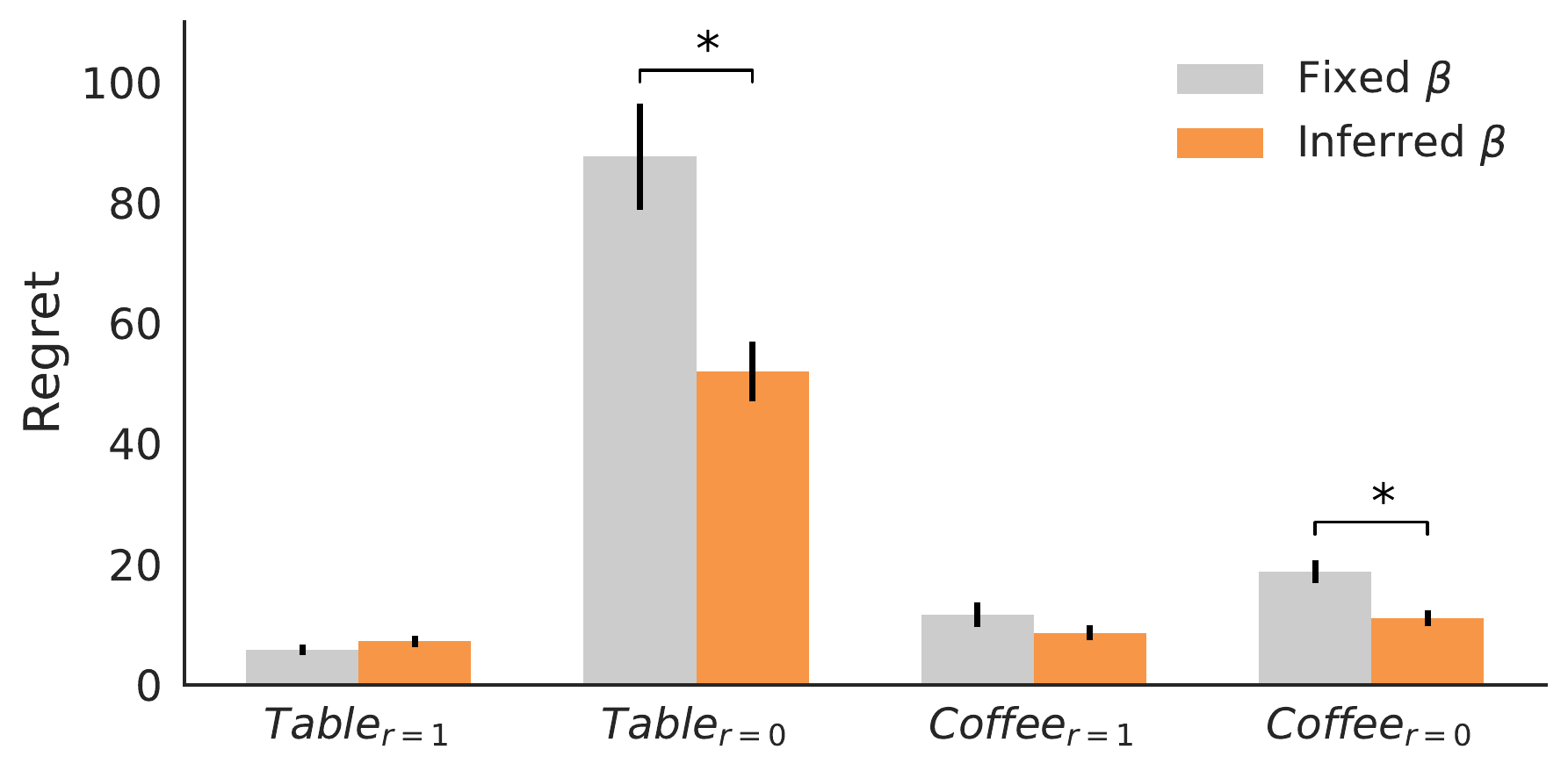}\\
\caption{Regret averaged across subjects.}
 \label{fig:regret}
\end{subfigure}
\begin{subfigure}[b]{0.5\textwidth}
\centering
\includegraphics[height=.5\textwidth]{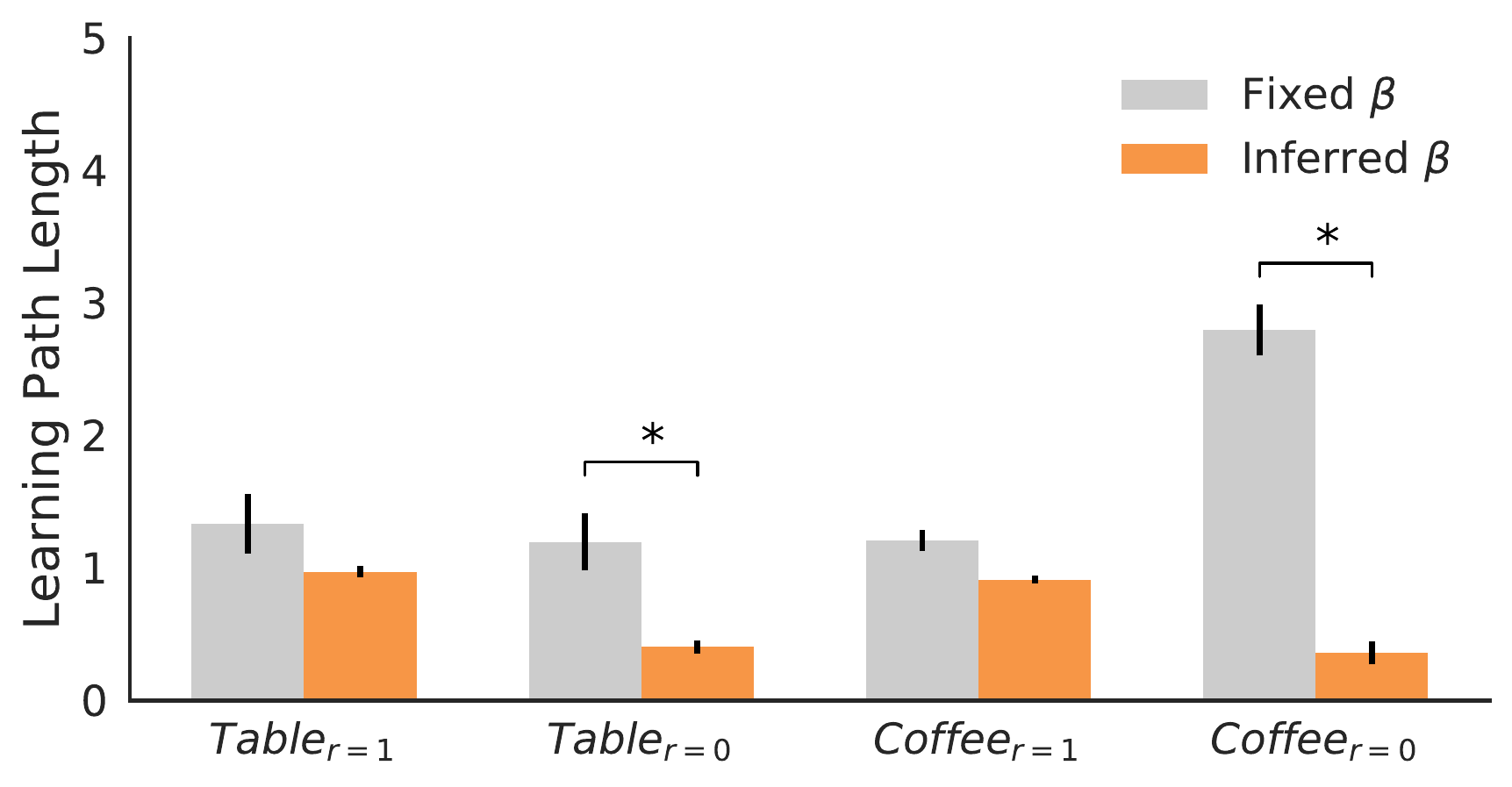}\\
\caption{$\hat{\theta}$ learning path length averaged across subjects.}
 \label{fig:weight_path}
\end{subfigure}
\vspace{0.5mm}
\caption{Comparison of regret and length of $\hat{\theta}$ learning path through weight space over time (lower is better).
}
\label{fig:delta_cost}
\end{figure}

\subsection{Analysis}

\noindent\textbf{Objective.}
We ran a repeated-measures factorial ANOVA with learning strategy and relevance as factors for the regret. We found a significant main effect for the method ($F(1,187)=7.8, p=0.0058$), and a significant interaction effect ($F(1,187)=6.77, p=0.0101$). We ran a post-hoc analysis with Tukey HSD corrections for multiple comparisons to analyze this effect, and found that it supported our hypotheses. On tasks where corrections were not relevant, our method had significantly lower regret ($p=0.001$); on tasks where corrections were relevant, there was no significant difference ($p=0.9991$). Figure \ref{fig:regret} plots the regret per task, and indeed our method was not inferior on tasks 1 and 3, and significantly better on tasks 2 and 4. We found analogous results on an ANOVA for the difference in learned weights from the optimal ones over time (see Figure \ref{fig:weight_path} for details). 

\noindent\textbf{Subjective.}
We ran a repeated measures ANOVA on the results of our participant survey. We find that our method is not significantly different from the baseline in terms of task completion ($F(1,7)=0.88,p=0.348$) and task understanding ($F(1,7)=0.55, p=0.46$), which supports hypothesis H3. At the same time, participants significantly preferred our method in terms of reducing unintended learning ($F(1,7)=9.15,p=0.0046$), which supports out final hypothesis, H4.

{\renewcommand{\arraystretch}{1.5}
\begin{table}[H]
 \centering
 \adjustbox{max width=\textwidth}{
 \begin{tabular}{|c|l|c|c|c|}
 \hline
 & \multicolumn{1}{c|}{\textbf{Questions}} & \multicolumn{1}{c|}{\textbf{Cronbach's $\alpha$}} & \multicolumn{1}{c|}{\textbf{F-Ratio}} & \multicolumn{1}{c|}{\textbf{p-value}} \\
 \hline
 \parbox[t]{2mm}{\multirow{2}{*}{\rotatebox[origin=c]{90}{\textbf{task}}}} 
 & The robot accomplished the task in the way I wanted. & \multirow{2}{*}{0.94} & \multirow{2}{*}{0.88} & \multirow{2}{*}{0.348}\\
 & The robot was NOT able to complete the task correctly.  &&&\\
 \hline
 \parbox[t]{2mm}{\multirow{2}{*}{\rotatebox[origin=c]{90}{\textbf{undst.}}}} 
 & I felt the robot understood how I wanted the task done. & \multirow{2}{*}{0.95} & \multirow{2}{*}{0.55} & \multirow{2}{*}{0.46}\\
 &  I felt the robot did NOT know how I wanted the task done. &&&\\
 \hline
 \parbox[t]{2mm}{\multirow{4}{*}{\rotatebox[origin=c]{90}{\textbf{unintend}}}} 
 & I had to undo corrections that I gave the robot. &&&\\
 & The robot wrongly updated its understanding about aspects of the task I did not want to change.  & \multirow{2}{*}{0.91} & \multirow{2}{*}{9.15} & \multirow{2}{*}{\textbf{0.0046}}\\
 & After I adjusted the robot, it continued to do the other parts of the task correctly.  &&&\\
 & After I adjusted the robot, it incorrectly updated parts of the task that were already correct. &&&\\
 \hline
 \end{tabular}
 }
 \strut 
 \label{table}
 \caption{Results of ANOVA on subjective metrics collected from a 7-point Likert-scale survey.} \label{tab:likert}
 \vspace{-5mm}
 \end{table}}

%% file: conclusion.tex
\section{Discussion}

\noindent\textbf{Summary.} We propose that robots should learn conservatively from human input by reasoning in real time about how rational the human's correction is for the robot's hypothesis space. Through both objective and subjective metrics, we show that this results in significantly reduced unintended learning when the human acts irrelevantly, and preserves performance when the actions are relevant.

\noindent\textbf{Limitations.} Our method is a first step in handling misspecified objectives by decreasing learning when an input is not understood. Ideally, the robot should both detect when features fall outside of its hypothesis space and identify them to continually augment its model.
Instead of using a number of hand-designed features, the robot could use data-driven techniques, such as function approximators, to obtain richer models. However, learning from a single correction in general becomes problematic due to the very limited training set compared to the space of possible feature misspecifications.

%% file: appendix.tex
\section{Appendix}







\subsection{Laplace Approximation of Denominator in Equation \ref{eq:bayesnetb_model}}\label{app:laplace}
Let the cost function in the denominator be denoted by $C(\bar{u}_H) = \beta( \|\bar{u}_H\|^2+ \lambda\|\Phi(\bar{\xi}_H) - \Phi(\xi_H)\|^2 )$ for an observed, $\Phi(\xi_H)$. First, our cost function can be approximated to quadratic order by computing a second order Taylor series approximation about the optimal human action $u^*_H$ (obtained via the constrained optimization in \ref{opt:optimal_uH}):
$$
C(\bar{u}_H) \approx C(u^*_H) + \nabla C(u^*_H)^{\top}(\bar{u}_H - u^*_H) + \frac{1}{2}(\bar{u}_H - u^*_H)^{\top}\nabla^2 C(u^*_H)(\bar{u}_H - u^*_H)\enspace.
$$
Since $\nabla C(\bar{u}_H)$ has a global minimum at $u^*_H$ then $\nabla C(u^*_H) = 0$ and the denominator of Equation \ref{eq:bayesnetb_model} can be rewritten as:
$$
\int e^{-C(\bar{u}_H)} d\bar{u}_H \approx e^{-C(u^*_H)}\int e^{-\frac{1}{2}(\bar{u}_H - u^*_H)\nabla^2 C(u^*_H)(\bar{u}_H - u^*_H)} d\bar{u}_H \enspace.
$$

Since $\nabla^2 C(u^*_H) > 0$ for $u_H^* \neq 0$, the integral is in Gaussian form, which admits a closed form solution:
$$
\int e^{-C(\bar{u}_H)} d\bar{u}_H \approx e^{ -C(u^*_H)} \sqrt{\frac{2\pi^k}{\beta^k|H_{u^*_H}|}} \enspace,
$$

where $H_{u^*_H} = \nabla^2 C(u^*_H)$ denotes the Hessian of the cost function at $u^*_H$. Replacing $C(\bar{u}_H)$ with the expanded cost function, we arrive at the final approximation of the observation model:
$$
P(u_H \mid \beta, \Phi(\xi_H), \xi_R) \approx \frac{e^{-\beta(  \|u_H\|^2)}}{ e^{-\beta( \|{u_H^*}\|^2 + \lambda\|\Phi({\xi_H^*}) - \Phi(\xi_H)\|^2)}}\sqrt{\frac{\beta^k|H_{u^*_H}|}{2\pi^k}}\enspace.
$$


%% file: corl_2018.bbl
\begin{thebibliography}{20}
\providecommand{\natexlab}[1]{#1}
\providecommand{\url}[1]{\texttt{#1}}
\expandafter\ifx\csname urlstyle\endcsname\relax
  \providecommand{\doi}[1]{doi: #1}\else
  \providecommand{\doi}{doi: \begingroup \urlstyle{rm}\Url}\fi

\bibitem[Abbeel and Ng(2004)]{abbeel2004apprenticeship}
P.~Abbeel and A.~Y. Ng.
\newblock Apprenticeship learning via inverse reinforcement learning.
\newblock In \emph{Machine Learning (ICML), International Conference on}. ACM,
  2004.

\bibitem[Osa et~al.(2018)Osa, Pajarinen, Neumann, Bagnell, Abbeel, Peters,
  et~al.]{osa2018algorithmic}
T.~Osa, J.~Pajarinen, G.~Neumann, J.~A. Bagnell, P.~Abbeel, J.~Peters, et~al.
\newblock An algorithmic perspective on imitation learning.
\newblock \emph{Foundations and Trends in Robotics}, 7\penalty0 (1-2):\penalty0
  1--179, 2018.

\bibitem[Javdani et~al.(2015)Javdani, Srinivasa, and
  Bagnell]{javdani2015shared}
S.~Javdani, S.~S. Srinivasa, and J.~A. Bagnell.
\newblock Shared autonomy via hindsight optimization.
\newblock \emph{arXiv preprint arXiv:1503.07619}, 2015.

\bibitem[Jain et~al.(2015)Jain, Sharma, Joachims, and Saxena]{jain2015learning}
A.~Jain, S.~Sharma, T.~Joachims, and A.~Saxena.
\newblock Learning preferences for manipulation tasks from online coactive
  feedback.
\newblock \emph{The International Journal of Robotics Research}, 34\penalty0
  (10):\penalty0 1296--1313, 2015.

\bibitem[Bajcsy et~al.(2017)Bajcsy, Losey, O'Malley, and Dragan]{CORL}
A.~Bajcsy, D.~P. Losey, M.~K. O'Malley, and A.~D. Dragan.
\newblock Learning robot objectives from physical human interaction.
\newblock In \emph{CoRL}, 2017.

\bibitem[Christiano et~al.(2017)Christiano, Leike, B.~Brown, Martic, Legg, and
  Amodei]{christiano2017preferences}
P.~Christiano, J.~Leike, T.~B.~Brown, M.~Martic, S.~Legg, and D.~Amodei.
\newblock Deep reinforcement learning from human preferences.
\newblock 06 2017.

\bibitem[Fu et~al.()Fu, Singh, Ghosh, Yang, and Levine]{fu2018variational}
J.~Fu, A.~Singh, D.~Ghosh, L.~Yang, and S.~Levine.
\newblock Variational inverse control with events: A general framework for
  data-driven reward definition.
\newblock \emph{arXiv preprint}, arXiv:1805.11686.

\bibitem[Hadfield-Menell et~al.(2017)Hadfield-Menell, Milli, Abbeel, Russell,
  and Dragan]{HadfieldMenell2017InverseRD}
D.~Hadfield-Menell, S.~Milli, P.~Abbeel, S.~J. Russell, and A.~D. Dragan.
\newblock Inverse reward design.
\newblock In \emph{NIPS}, 2017.

\bibitem[Kalman(1964)]{Kalman1964inverse}
R.~E. Kalman.
\newblock {When Is a Linear Control System Optimal?}
\newblock \emph{Journal of Basic Engineering}, 86\penalty0 (1):\penalty0
  51--60, mar 1964.
\newblock ISSN 0098-2202.
\newblock URL \url{http://dx.doi.org/10.1115/1.3653115}.

\bibitem[Ng and Russell(2000)]{Ng2000inverse}
A.~Ng and S.~Russell.
\newblock {Algorithms for inverse reinforcement learning}.
\newblock \emph{International Conference on Machine Learning (ICML)},
  0:\penalty0 663--670, 2000.
\newblock ISSN 00029645.
\newblock \doi{10.2460/ajvr.67.2.323}.
\newblock URL
  \url{http://www-cs.stanford.edu/people/ang/papers/icml00-irl.pdf}.

\bibitem[Ramachandran and Amir()]{ramachandran2007bayesian}
D.~Ramachandran and E.~Amir.
\newblock Bayesian inverse reinforcement learning.
\newblock \emph{Urbana}, 51\penalty0 (61801):\penalty0 1--4.

\bibitem[Milli et~al.(2017)Milli, Hadfield-Menell, Dragan, and
  Russell]{milli2017should}
S.~Milli, D.~Hadfield-Menell, A.~Dragan, and S.~Russell.
\newblock Should robots be obedient?
\newblock \emph{arXiv preprint arXiv:1705.09990}, 2017.

\bibitem[Bajcsy et~al.(2018)Bajcsy, Losey, O'Malley, and Dragan]{HRI}
A.~Bajcsy, D.~P. Losey, M.~K. O'Malley, and A.~D. Dragan.
\newblock Learning from physical human corrections, one feature at a time.
\newblock In \emph{Proceedings of the 2018 ACM/IEEE International Conference on
  Human-Robot Interaction}, HRI '18, pages 141--149, New York, NY, USA, 2018.
  ACM.
\newblock ISBN 978-1-4503-4953-6.
\newblock \doi{10.1145/3171221.3171267}.
\newblock URL \url{http://doi.acm.org/10.1145/3171221.3171267}.

\bibitem[Fisac et~al.(2018)Fisac, Bajcsy, Herbert, Fridovich-Keil, Wang,
  Tomlin, and Dragan]{fisac2018probabilistically}
J.~F. Fisac, A.~Bajcsy, S.~L. Herbert, D.~Fridovich-Keil, S.~Wang, C.~J.
  Tomlin, and A.~D. Dragan.
\newblock Probabilistically safe robot planning with confidence-based human
  predictions.
\newblock \emph{Robotics: Science and Systems (RSS)}, 2018.

\bibitem[Ziebart et~al.(2008)Ziebart, Maas, Bagnell, and Dey]{maxent}
B.~D. Ziebart, A.~Maas, J.~A. Bagnell, and A.~K. Dey.
\newblock Maximum entropy inverse reinforcement learning.
\newblock In \emph{Proceedings of the 23rd National Conference on Artificial
  Intelligence - Volume 3}, AAAI'08, pages 1433--1438. AAAI Press, 2008.
\newblock ISBN 978-1-57735-368-3.
\newblock URL \url{http://dl.acm.org/citation.cfm?id=1620270.1620297}.

\bibitem[Von~Neumann and Morgenstern(1945)]{von1945theory}
J.~Von~Neumann and O.~Morgenstern.
\newblock \emph{Theory of games and economic behavior}.
\newblock Princeton University Press Princeton, NJ, 1945.

\bibitem[Baker et~al.(2007)Baker, Tenenbaum, and Saxe]{baker2007goal}
C.~L. Baker, J.~B. Tenenbaum, and R.~R. Saxe.
\newblock Goal inference as inverse planning.
\newblock In \emph{Proceedings of the Annual Meeting of the Cognitive Science
  Society}, volume~29, 2007.

\bibitem[Dragan et~al.(2015)Dragan, Muelling, Bagnell, and
  Srinivasa]{deformation}
A.~D. Dragan, K.~Muelling, J.~A. Bagnell, and S.~S. Srinivasa.
\newblock Movement primitives via optimization.
\newblock In \emph{2015 IEEE International Conference on Robotics and
  Automation (ICRA)}, pages 2339--2346, May 2015.
\newblock \doi{10.1109/ICRA.2015.7139510}.

\bibitem[Schulman et~al.()Schulman, Ho, Lee, Awwal, Bradlow, and
  Abbeel]{trajopt}
J.~Schulman, J.~Ho, A.~Lee, I.~Awwal, H.~Bradlow, and P.~Abbeel.
\newblock Finding locally optimal, collision-free trajectories with sequential
  convex optimization.

\bibitem[Hogan(1985)]{impedance}
N.~Hogan.
\newblock Impedance control: An approach to manipulation: Part
  ii--implementation.
\newblock \emph{Journal of Dynamic Systems, Measurement, and Control},
  107\penalty0 (1):\penalty0 8--16, Mar 1985.
\newblock ISSN 0022-0434.
\newblock \doi{10.1115/1.3140713}.
\newblock URL \url{http://dx.doi.org/10.1115/1.3140713}.

\end{thebibliography}
